\documentclass[11pt]{article}

\usepackage[final]{acl}

\usepackage{times}
\usepackage{latexsym}
\usepackage{float}

\usepackage[T1]{fontenc}

\usepackage[utf8]{inputenc}

\usepackage{microtype}

\usepackage{inconsolata}

\usepackage{graphicx}
\usepackage{CJKutf8}
\usepackage{enumitem}
\usepackage{booktabs}
\usepackage{xpinyin}
\usepackage{multirow} 
\usepackage{tcolorbox}
\usepackage{dashrule}
\usepackage{array}

\usepackage{amsmath,amsfonts,bm}

\def\1{\bm{1}}

\def\vx{{\bm{x}}}
\def\vy{{\bm{y}}}

\DeclareMathAlphabet{\mathsfit}{\encodingdefault}{\sfdefault}{m}{sl}
\SetMathAlphabet{\mathsfit}{bold}{\encodingdefault}{\sfdefault}{bx}{n}

\def\gD{{\mathcal{D}}}

\def\gL{{\mathcal{L}}}

\title{\textsc{GUIDE}: Generative Unsupervised Chinese Query Correction via Phonetic and Visual Shared-ID Encoding}

  \author{
        \textbf{Lei Yang\textsuperscript{1}}\thanks{\ Equal contribution.
  \ $^{\dagger}$Corresponding author.},
        \textbf{Binbin Huang\textsuperscript{1,2}}\footnotemark[1],
        \textbf{Jiwei Tan\textsuperscript{1}}$^{\dagger}$,
        \textbf{Xuhui Sui\textsuperscript{1}},
        \textbf{Chang Tu\textsuperscript{1}},
        \textbf{Yi Wang\textsuperscript{1}},
        \textbf{Han Li\textsuperscript{1}}
        \\
        \\
        \textsuperscript{1}Kuaishou Technology,
        \textsuperscript{2}School of Data Science, Fudan University
        \\
        \small{
          \textbf{Correspondence:} \href{mailto:email@domain}{chentiancheng03@kuaishou.com}
        }
    }
  
\begin{document}
\maketitle

\begin{abstract}

Chinese query correction (CQC) is important for search and query recommendation on content platforms, but supervised methods rely on large annotated correction pairs that are costly to maintain as query vocabularies evolve. 
Unsupervised correction with language models is attractive, yet in the short-query setting, unconstrained generation often over-corrects ambiguous inputs toward high-frequency phrases, causing intent drift. 
We propose \textsc{GUIDE}, a generative unsupervised framework for CQC based on a confuse-then-clarify paradigm. 
\textsc{GUIDE} encodes phonetically or visually confusable characters with shared-IDs and reconstructs the original query with an encoder--decoder architecture, which constrains correction to plausible confusion neighborhoods while learning from unlabeled query streams. 
A time-decayed, query-frequency-weighted objective further supports adaptation to rapidly changing query vocabularies. 
Experiments on \textit{QSpell 250K} and \textit{KwaiSearch}, a large-scale industrial dataset from Kuaishou, show that \textsc{GUIDE} consistently outperforms strong baselines, while online A/B testing further confirms gains in correction quality and downstream engagement.

\end{abstract}

\section{Introduction}
\label{sec:introduction}

Search queries directly express user intent and determine what content is retrieved and surfaced.
On large content platforms such as TikTok and Kuaishou, query recommendation modules---including search suggestions, autocomplete, and related-query recommendation---are often built from historical search logs and user interactions \cite{bacciu-etal-2024-gqr}. 
As a result, misspellings and non-canonical variants in query streams can be repeatedly exposed, clicked, and re-mined, causing noisy forms to propagate through downstream retrieval, ranking, and recommendation pipelines \cite{gao-etal-2010-large,sun-etal-2012-fast}. 
This makes \emph{Chinese query correction} (CQC) an important problem in industrial search systems.

Unlike the well-studied sentence-level Chinese spelling correction (CSC), CQC is substantially harder in practice due to several query-specific constraints~\cite{yang2023chinese,su-etal-2025-racqc}:
\begin{CJK*}{UTF8}{gkai}
(1) \textit{Short Context}: queries provide little context, making it hard to decide whether to edit and what is intended; 
(2) \textit{Fast Shift}: vocabularies change rapidly with new memes, influencers, and emerging entities; 
(3) \textit{Creative Use}: homophones can be intentional puns---e.g., ``钱途'' (lit.\ ``money road'', i.e.,\ ``money-
  making prospects'') puns on its homophone ``前途'' (\emph{qiántú}, ``future prospects'')---so different is not always wrong; and
(4) \textit{Few Labels}: large-scale query annotations are expensive, and even millions of labeled pairs cover only a small fraction of real errors.
\end{CJK*} These constraints make CQC fundamentally a problem of controlled editing: the system must decide not only what to correct, but also when to correct and how far it may deviate from the original expression.

Existing approaches only partially address this challenge. 
Supervised correctors learn a correction policy from labeled pairs \cite{zhang-etal-2020-spelling,xu-etal-2021-read}, but collecting and refreshing such annotations is expensive under fast vocabulary shift. 
Weakly supervised methods typically construct pseudo pairs through synthetic corruption \cite{liu-etal-2021-plome,li-2022-uchecker}, which introduces a hand-designed noise process that may not match real query errors. 
Training-free Language Model-based correction is attractive \cite{hong-etal-2019-faspell,zhou-etal-2024-simple}, but unconstrained or weakly constrained decoding often over-corrects short ambiguous queries, introduces unnecessary edits, or fails to preserve the original query length \cite{li2023effectiveness,liu-etal-2024-relm}.

These limitations suggest that effective unsupervised CQC should not rely on free-form rewriting. 
Instead, correction should stay within plausible confusion neighborhoods induced by realistic input errors and shaped by the language's input system. 
Based on this intuition, we propose \textsc{GUIDE}, a generative unsupervised framework built on a confuse-then-clarify paradigm. 
The core idea is to map confusable characters into shared-IDs and then train an encoder--decoder model to reconstruct the original character sequence. 
This shared-ID reconstruction objective turns character confusion structure into a learning signal: the encoder is encouraged to tolerate realistic ambiguity, while the decoder is forced to clarify it back into a concrete query. 
As a result, \textsc{GUIDE} learns controlled correction from unlabeled query streams without requiring manually annotated correction pairs.

For Chinese, these neighborhoods are naturally instantiated by the two dominant query errors: \textit{sound-alike} substitutions and, to a lesser extent, \textit{look-alike} ones \cite{liu-etal-2010-visually,wu-etal-2013-chinese}. 
These error patterns are closely tied to Chinese Input Method Editors (IMEs), such as pinyin and handwriting. 
Accordingly, we implement \textsc{GUIDE} with two complementary clustering strategies: phonetic shared-ID clustering based on pinyin confusability, and visual shared-ID clustering based on image-based character similarity.
On top of these clustered inputs, we train an encoder--decoder Transformer with a time-decayed, query-frequency-weighted objective so that the model can continually adapt to recent and frequent queries in evolving search traffic. Our contributions are as follows:
\begin{itemize}[leftmargin=1.0em,itemsep=0.1em,topsep=0.0em,parsep=0.0cm]
    \item We propose \textsc{GUIDE}, a \textbf{G}enerative \textbf{U}nsupervised framework for Chinese query correction based on a confuse-then-clarify paradigm, where phonetic and visual shared-\textbf{ID} \textbf{E}ncoding provides an explicit inductive bias for controlled correction under weak query context.
    \item We show how shared-ID reconstruction enables learning from unlabeled query streams without error--correction pairs.
    \item We release \textit{KwaiSearch}, a real-world search-log dataset from the Kuaishou platform, and show strong results on both \textit{QSpell 250K} and \textit{KwaiSearch}, supported by online A/B testing.
\end{itemize}

\vspace{-0.5em}
\section{Methodology}
\label{sec:methodology}

\subsection{Problem Formulation and Overview}

Let $\gD=\{(\vx_i,t_i)\}_{i=1}^{N}$ denote a training set of $N$ Chinese character queries collected from logs, where $\vx_i=\{x_{i,1},x_{i,2},\ldots,x_{i,m_i}\}$ is the observed (possibly incorrect) query at time $t_i$. 
The CQC task maps an input query $\vx_i$ to a corrected query $\vy_i=\{y_{i,1},y_{i,2},\ldots,y_{i,m_i}\}$, where $\vy_i$ has the same length as $\vx_i$ and differs only at erroneous character positions. 
In CQC, errors often arise from phonetically or visually similar candidates, with representative examples shown in Table~\ref{tab:similarity_examples}.

\begin{CJK*}{UTF8}{gkai} 
\begin{table}[t]
\centering
\begin{tabular}{p{1.7cm}p{5.2cm}}
\toprule
\multicolumn{2}{@{}l}{\textbf{Phonetically Similar Case}} \\
\midrule
{Input} & 土坡上的\textcolor{red}{苟(\pinyin{gou3})}尾草。\\
{PSC} & 苟 (\pinyin{gou3})\enspace\textcolor{blue}{狗 (\pinyin{gou3})}\enspace够 (\pinyin{gou4})\enspace钩 (\pinyin{gou1})\\
\toprule
\multicolumn{2}{@{}l}{\textbf{Visually Similar Case}} \\
\midrule
{Input} & 土坡上的\textcolor{red}{狍(\pinyin{pao2})}尾草。\\
{VSC} & 狍(\pinyin{pao2})\enspace\textcolor{blue}{狗(\pinyin{gou3})}\enspace 句(\pinyin{ju4})\enspace 拘(\pinyin{ju1})\\
\midrule
{Correct} & 土坡上的\textcolor{blue}{狗(\pinyin{gou3})}尾草。\\
{Translation} & Bristlegrass on a Slope.\\
\bottomrule
\end{tabular}
\caption{Examples of phonetically similar candidates (PSC) and visually similar candidates (VSC) in Chinese spelling errors.
  \textcolor{red}{Misspelled} characters are marked in red, while the \textcolor{blue}{correct} forms are indicated in blue. Phonetic
  errors are same-syllable confusions (often differing only in tone) from pinyin typing; visual errors are look-alike characters that
  typically share a radical/component.}\vspace{-1.5em}
\label{tab:similarity_examples}
\end{table}
\end{CJK*}

\begin{figure*}[t]
  \includegraphics[width=0.49\linewidth]{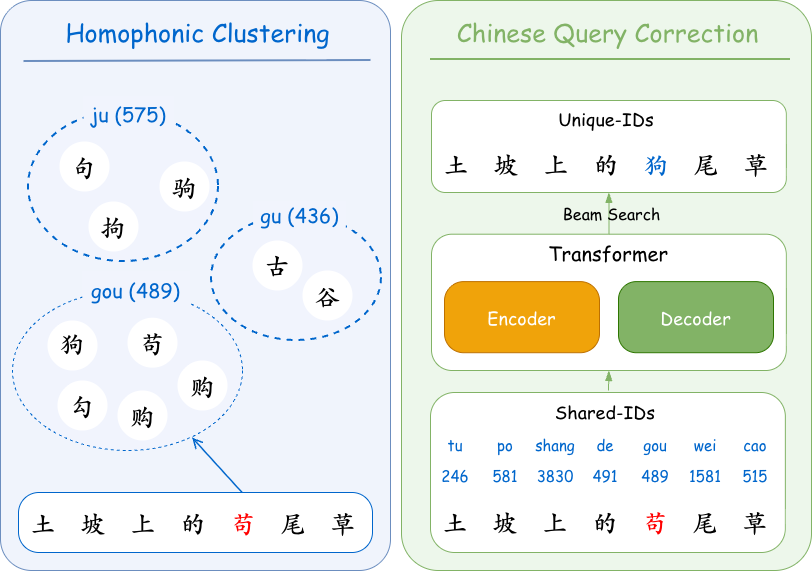} \hfill
  \includegraphics[width=0.49\linewidth]{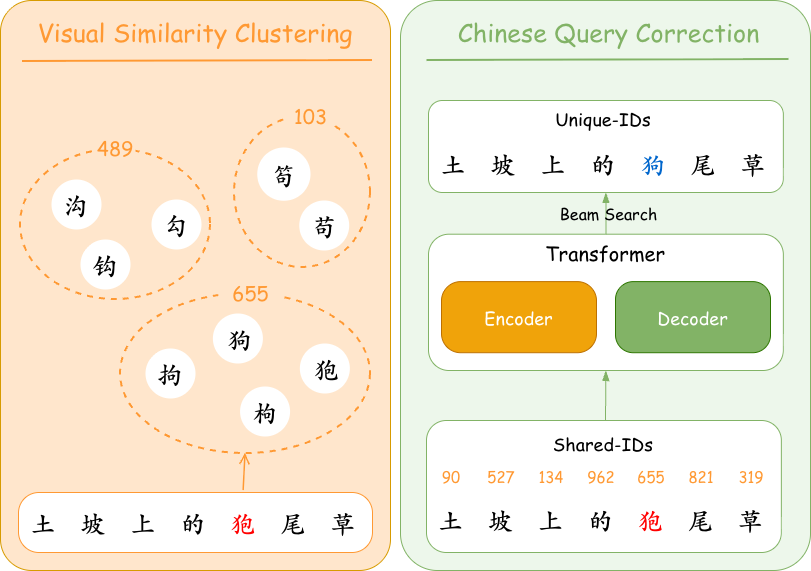}
\vspace{-0.2em}
\caption{Overview of \textsc{GUIDE} with two clustering-based correction models: the left part shows the homophonic model, and the right part shows the visual-similarity model, based on pronunciation and character shape, respectively.\vspace{-1.7em}}
  \label{fig:GUIDE-structure}
\end{figure*}

To avoid unintended rewrites under weak query context, \textsc{GUIDE} constrains correction through character clustering and sequence reconstruction. 
As shown in Figure~\ref{fig:GUIDE-structure}, the framework consists of two core stages: (1) \textit{character clustering}, which maps an input query into a phonetic or visual shared-ID sequence, and (2) \textit{Chinese query correction}, which reconstructs the original character sequence with an encoder--decoder Transformer.

\subsection{Character Clustering}
\label{sec:character_clustering}

Character clustering groups Chinese characters by phonetic or visual similarity and maps each character to a shared ID. 
Formally, let $g(\cdot)$ be a clustering-based mapping that converts a query $\vx_i$ into a shared-ID sequence
$\tilde{\vx}_i=g(\vx_i)=\{\tilde{x}_{i,1},\ldots,\tilde{x}_{i,m_i}\}$,
where multiple confusable characters may share the same ID. 
This mapping is intentionally lossy: it removes fine-grained character identity at the encoder side, making the input more tolerant to typical user errors and restricting edits to plausible confusions.
To cover dominant error types in practice, especially phonetic confusions and visually similar confusions, we design two complementary ways to construct $g(\cdot)$.
\vspace{-.5em}
\paragraph{Homophonic Clustering.}
Most query typos on content platforms are sound-alike confusions caused by pinyin input. 
We therefore build a pinyin-based clustering strategy that maps each character to a pronunciation key \emph{without tones}, and assigns the same shared ID to characters with the same tone--stripped pinyin. 
For polyphonic characters, we choose the most frequent tone--stripped pronunciation in query logs to determine the ID, which aligns the clusters with common user inputs and captures frequent phonetic confusions.

\vspace{-.5em}
\paragraph{Visual Similarity Clustering.}
To handle another major error type---look-alike confusions---we cluster characters by visual similarity using image-based character representations. We render each character as an image and encode it with a Vision Transformer (ViT)~\cite{dosovitskiy2020image} to obtain feature vectors, then cluster characters using cosine similarity between these representations. We use a threshold~$\tau$ to determine whether two characters should be grouped into the same cluster, thereby controlling the granularity of visual neighborhoods.

\subsection{Chinese Query Correction}
\label{sec:chinese_query_correction}

We perform Chinese query correction with an encoder--decoder Transformer, where the encoder consumes shared-ID inputs and the decoder predicts the original character IDs.
In the encoder stage, input query characters are mapped to shared-IDs, enabling the model to learn robust representations of error variants from large-scale, mostly error-free text data. The decoder then maps the shared-ID sequence back to the corresponding original character IDs. During inference, we decode the output sequence using beam search.

In practice, we train two separate correction models based on the two clustering strategies, namely the homophonic model and the visual-similarity model. The homophonic and visual-similarity models can be applied independently or combined. For deployment, we use a simple dynamic fusion without an explicit error-type classifier: for the same query, we feed the phonetic shared-ID sequence to the homophonic model and the visual shared-ID sequence to the visual-similarity model, and obtain decoder scores (token logits) for beam candidates from both models. We then compare the candidates from the two models and keep the one with the smaller score gap to the original query (i.e., the smaller change in normalized sequence log-likelihood relative to the original query), which helps prevent over-correction while still benefiting from beam search for plausible alternatives.

\subsection{Training Objective with Time-decayed Frequency Reweighting}
Given the shared-ID sequence $\tilde{\vx}_i$ obtained by the clustering-based mapping in Section~\ref{sec:character_clustering}, 
we then train \textsc{GUIDE} to reconstruct the original query $\vx_i=\{x_{i,1},\ldots,x_{i,m_i}\}$ with a time-decayed frequency-reweighted negative log-likelihood:
\begin{equation}
\gL = -\frac{1}{N}\sum_{i=1}^{N} w_i \sum_{j=1}^{m_i}\log p_\theta(x_{i,j}\mid \tilde{\vx}_i, x_{i,<j}),
\label{eq:time-decayed-nll}
\end{equation}
where $w_i = \log(1 + c_i)\cdot \exp(-\lambda\cdot \Delta t_i)$ is a weight that depends on the query search count $c_i$ and freshness $\Delta t_i$. This reweighting shifts learning focus toward frequent and recent queries while preserving the shared-ID reconstruction objective.

\section{Experiments}
\label{sec:experiments}

\begin{table*}[t]
\centering
\setlength{\tabcolsep}{3pt}
\begin{tabular}{
 c|c|p{3.4cm}
 |c@{\hspace{12pt}}c@{\hspace{10pt}}c
 |c@{\hspace{12pt}}c@{\hspace{10pt}}c
}
\toprule
\multirow{2}{*}{\parbox{0.9cm}{\centering \textbf{}}}
& \multirow{2}{*}{\parbox{2.5cm}{\centering \textbf{Method}}}
& \multirow{2}{*}{\parbox{3.4cm}{\centering \textbf{Model Size/Setting}}}
& \multicolumn{3}{c|}{\textit{QSpell 250K}}
& \multicolumn{3}{c}{\textit{KwaiSearch}} \\
&  &  &
\textbf{Prec.} & \textbf{Rec.} & \textbf{F1}
& \textbf{Prec.} & \textbf{Rec.} & \textbf{F1} \\
\midrule

\multirow{2}{*}{\rotatebox{90}{\textbf{Sup.}}} &
\textsc{BERT} & \multicolumn{1}{c|}{---} &
0.1891& 0.4385& 0.2642& 0.0624& 0.2330& 0.0984\\
\cmidrule(lr){2-9}
& \textsc{Masked-FT} & \multicolumn{1}{c|}{---} &
0.2931& 0.6247& 0.3990& 0.1603& 0.5734& 0.2505\\

\midrule

\multirow{12}{*}{\rotatebox{90}{\textbf{Unsup.}}} &
\multirow{3}{*}{\textsc{Simple-CSC}} &
Qwen3-0.6B &
0.3224& 0.6693& 0.4352& 0.1450& 0.6174& 0.2348\\
& & Qwen3-1.7B &
0.3282& \textbf{0.7379}& 0.4543& 0.1554& 0.7156& 0.2553\\
& & Qwen3-4B &
0.3617& 0.6950& 0.4757& 0.1732& 0.6454& 0.2730\\

\cmidrule(lr){2-9}

& \multirow{6}{*}{\textsc{LLM-ICL}} &
Qwen3-0.6B\hfill 0-shot &
0.0365& 0.0954& 0.0528& 0.0322& 0.1290& 0.0516\\
& & Qwen3-0.6B\hfill 10-shot &
0.0352& 0.1495& 0.0569& 0.0305& 0.4424& 0.0570\\
& & Qwen3-1.7B\hfill 0-shot &
0.0786& 0.3046& 0.1249& 0.0251& 0.1859& 0.0442\\
& & Qwen3-1.7B\hfill 10-shot &
0.0975& 0.5375& 0.1650& 0.0322& 0.5010& 0.0601\\
& & Qwen3-4B\hfill 0-shot &
0.1645& 0.2620& 0.2021& 0.0664& 0.1937& 0.0989\\
& & Qwen3-4B\hfill 10-shot &
0.2105& 0.5078& 0.2976& 0.0730& 0.3424& 0.1203\\

\cmidrule(lr){2-9}

& \multirow{3}{*}{\textbf{\textsc{GUIDE} (ours)}} &
3-layer\hfill Enc--Dec &
0.4244& 0.5341& 0.4730& 0.6738& 0.8476& 0.7508\\
& & 6-layer\hfill Enc--Dec &
\textbf{0.4367}& 0.5547& \textbf{0.4887}& \textbf{0.6821}& \textbf{0.8419}& \textbf{0.7536}\\
& & 12-layer\hfill Enc--Dec &
0.4321& 0.5471& 0.4829& 0.6749& 0.8364& 0.7474\\

\bottomrule
\end{tabular}
\caption{Main results on \textit{QSpell 250K} and \textit{KwaiSearch}. Sup./Unsup. denote supervised/unsupervised methods. Prec., Rec., and F1 denote precision, recall, and F1 score, respectively. \textbf{bold} indicates the best model for each dataset/metric.\vspace{-1.5em}}
\label{tab:exp_results}
\end{table*}

This section evaluates \textsc{GUIDE} for CQC with both offline and online studies. We report offline results on the public \textit{QSpell 250K} benchmark and our in-house dataset \textit{KwaiSearch}. We further validate \textsc{GUIDE} via an online A/B test on production traffic.

\subsection{Experimental Setup}

\paragraph{Datasets.}
We evaluate \textsc{GUIDE} on two datasets that better reflect the practical CQC setting on content platforms.
This differs from prior Chinese Spelling Correction (CSC) benchmarks (e.g., SIGHAN13/14/15 \cite{wu2013chinese, yu2014overview, tseng2015introduction}, Wang271K \cite{wang2018hybrid}, and LEMON \cite{wu-etal-2023-rethinking}), which are mostly sentence-oriented and less aligned with query noise patterns.
Specifically, we use (i) \textit{QSpell 250K} \cite{ye-etal-2025-qspell}, a public simplified-Chinese query correction benchmark with short inputs; we train \textsc{GUIDE} using only the provided corrected queries;
 and (ii) \textit{KwaiSearch}, our in-house dataset built from user search logs on the Kuaishou platform. It contains queries and short texts (e.g., titles and OCR-derived snippets) extracted from key search-result content, along with search-count and timestamp signals.\footnote{\ The dataset is
  available at \url{https://github.com/lyang05/GUIDE}.}
Table~\ref{tab:dataset_stats} summarizes the dataset statistics.

\begin{table}[t]
\centering
\small
\setlength{\tabcolsep}{4pt}
\begin{tabular}{@{}lcccc@{}}
\toprule
\textbf{Name} & \textbf{Usage} & \textbf{\#Sent} & \textbf{\#Error} & \textbf{Avg. Len.}\\
\midrule
\multirow{2}{*}{\textit{QSpell 250K}} & Train & 200K & 102K & 8.56 \\
                            & Test  & 50K  & 26K  & 8.58 \\
\midrule
\multirow{2}{*}{\textit{KwaiSearch}} & Train & 180M& Unknown & 9.67\\
& Test  & 30K& 15K& 7.35\\
\bottomrule
\end{tabular}
\caption{Statistics of all datasets. \#Sent, \#Error, and Avg. Len.\ denote the number of queries, error queries, and average query length, respectively.\vspace{-1.5em}}
\label{tab:dataset_stats}
\end{table}

\vspace{-.5em}
\paragraph{Evaluation Metrics.}
We report query-level Precision, Recall, and F1, which are standard in CSC. 

\vspace{-.5em}
\paragraph{Baseline Methods.}
We compare \textsc{GUIDE} with supervised and unsupervised baselines:
\begin{itemize}[noitemsep, parsep=0.5pt, partopsep=0pt, leftmargin=10pt]
    \item \textsc{BERT} \cite{devlin-etal-2019-bert}: a supervised correction baseline that fine-tunes BERT-base-Chinese to predict corrected characters in context.
    \item \textsc{Masked-FT} \cite{wu-etal-2023-rethinking}: a supervised fine-tuning strategy that masks non-error characters during training to reduce trivial copying and mitigate over-correction.
    \item \textsc{Simple-CSC} \cite{zhou-etal-2024-simple}: a training-free, prompt-free LLM baseline that treats the LLM as a left-to-right language model and corrects text by constrained decoding, using a minimal-distortion constraint (phonetic/visual similarity) to keep the output close to the input.
    \item \textsc{LLM-ICL}: a prompting baseline using the Qwen3 series \cite{yang2025qwen3}. We evaluate 0-shot and 10-shot settings on three model sizes (0.6B/1.7B/4B) with randomly sampled demonstrations.
\end{itemize}

\vspace{-1.5em}
\paragraph{Implementation Details.}
We instantiate \textsc{GUIDE} with three Transformer encoder--decoder sizes, using 3-, 6-, and 12-layer encoders and decoders, respectively. 
Unless otherwise specified, we use a single unified system that combines the homophonic and visual-similarity models. 
Since \textit{KwaiSearch} has no ground-truth labels, supervised baselines are trained on \textit{QSpell 250K} and evaluated on \textit{KwaiSearch}.
Detailed model configurations and training hyperparameters are provided in Appendix~\ref{app:training_settings}.






\begin{table}[t]
\centering
\small
\setlength{\tabcolsep}{3pt}
\resizebox{\columnwidth}{!}{%
\begin{tabular}{l|ccc|ccc}
\toprule
\multirow{2}{*}{Variant} 
& \multicolumn{3}{c|}{\textit{QSpell 250K}} 
& \multicolumn{3}{c}{\textit{KwaiSearch}} \\
& Prec. & Rec. & F1 & Prec. & Rec. & F1 \\
\midrule
Pho-only & 0.4107& 0.5230& 0.4628& 0.6661& 0.8433& 0.7443\\
Vis-only & 0.0807& \textbf{0.8152}& 0.1470& 0.1326& 0.7971& 0.2274\\
Pho+Vis  & \textbf{0.4244}& 0.5341& \textbf{0.4730}& \textbf{0.6738}& \textbf{0.8476}& \textbf{0.7508}\\
\bottomrule
\end{tabular}}
\caption{Comparison of clustering strategies in \textsc{GUIDE}. ``Pho'' and ``Vis'' denote the homophonic model and the visual-similarity model, respectively. ``Pho+Vis'' is the combined setting used in our main experiments.\vspace{-1.5em}}
\label{tab:ablation_homo_vis}
\end{table}

\subsection{Main Results}

Table~\ref{tab:exp_results} reports results on \textit{QSpell 250K} and \textit{KwaiSearch}. \textsc{GUIDE} achieves the best overall performance on both datasets, with especially large gains on \textit{KwaiSearch}. This suggests that constraining correction to phonetic/visual confusion neighborhoods is highly effective.
On \textit{QSpell 250K}, \textsc{Simple-CSC} trades precision for recall, reflecting more aggressive rewriting behavior. In query recommendation, however, false corrections are often more harmful than missed typos, so the stronger precision--recall balance of \textsc{GUIDE} is better aligned with deployment needs. In-context prompting remains consistently weaker overall.  Additional experimental results are provided in Appendix~\ref{app:additional_results}.

\subsection{Online A/B Testing}
\label{subsec:AB_testing}
We further validate \textsc{GUIDE} in an online A/B test on real search traffic.
Here we deploy the corrector at several high-traffic query entry points, where it corrects the candidate query set for recommendation: SUG search (search suggestions during typing), guess you search (pre-search query guesses surfaced to users), boxed-term search (search-box guesses of likely queries), and comment/bottom-bar guess search (lightweight guesses shown in comment or bottom-bar modules).
As the production baseline, we compare with the previous online correction method. Compared with this baseline, \textsc{GUIDE} substantially improves user-perceived quality: the misspelling rate drops by 80.1\% (from 2.01\% to 0.40\%). 
We also observe consistent efficiency gains across multiple entry points, including SUG search PV (+0.152\%), guess you search PV (+0.576\%), boxed-term search PV (+0.707\%), and comment/bottom-bar guess search PV (+0.986\%), together leading to a +0.122\% increase in overall search volume. Full experimental settings—including the baseline model, traffic allocation, test duration, evaluation protocol, and post-rollout monthly tracking—are provided in Appendix~\ref{app:ab_details}.

\section{Discussion}
\label{sec:discussion}
\vspace{-.2em}
\subsection{Are Homophonic and Visual-Similarity Models Complementary?}
\vspace{-.2em}
To verify whether \textsc{GUIDE} benefits from modeling both dominant error sources in Chinese queries, we compare three settings: (i) only the homophonic model, (ii) only the visual-similarity model, and (iii) their combination (our default setting). 
As shown in Table~\ref{tab:ablation_homo_vis}, Pho+Vis achieves the best F1 on both datasets, indicating that the two clustering strategies are complementary. Overall, the homophonic model plays a more dominant role, while the visual-similarity model provides additional gains by covering look-alike errors that are not well captured by pronunciation cues.


\begin{figure}[t]
    \centering
    \includegraphics[width=1\linewidth]{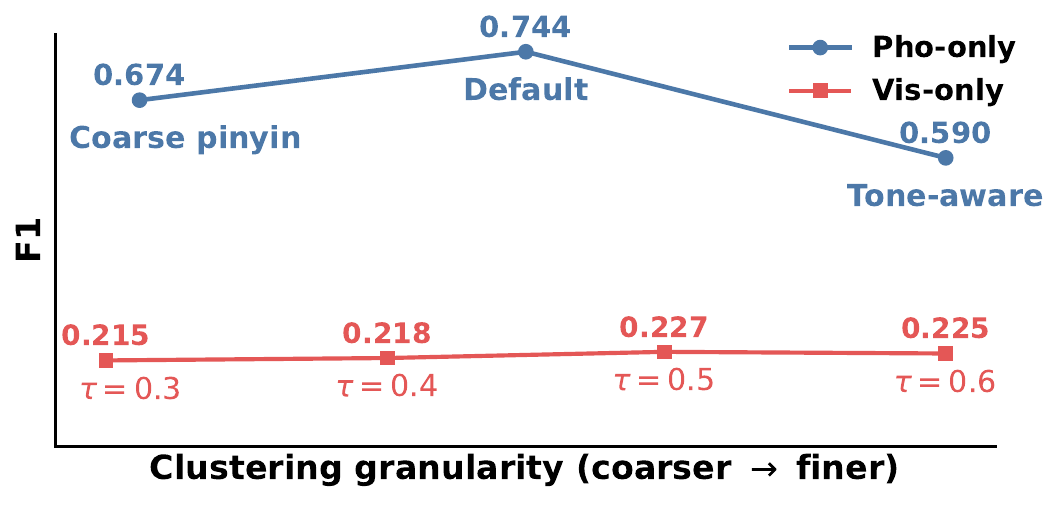}
    \vspace{-1.5em}
    \caption{Sensitivity of \textsc{GUIDE} to phonetic and visual clustering granularity on \textit{KwaiSearch}.}
    \label{fig:granularity_sensitivity}
    \vspace{-1.em}
\end{figure}

\vspace{-.25em}
\subsection{How Sensitive Is \textsc{GUIDE} to Clustering Granularity?}
\label{subsec:clustering_granularity_study}
\vspace{-.25em}
We examine the effect of clustering granularity on correction quality. 
For homophonic clustering, we compare the default tone-stripped setting with two variants: a coarser pinyin setting and a finer tone-aware setting.
The coarser setting merges common pinyin confusions that arise from regional accents in Mandarin and are directly reflected as typing
  errors in pinyin IMEs, including retroflex/non-retroflex initials (\{z, zh\}, \{c, ch\}, \{s, sh\}), nasal/lateral initials (\{n, l\}), labiodental confusions (\{h, f\}), and front/back nasal finals (\{en, eng\}, \{in, ing\}, \{an, ang\}). 
For visual clustering, we vary the threshold $\tau$ that controls neighborhood size. 
Figure~\ref{fig:granularity_sensitivity} summarizes the trends, while the full numerical results are provided in Table~\ref{tab:granularity_sensitivity}.
The default tone-stripped clustering gives the best overall F1 among phonetic variants, while $\tau=0.5$ performs best for visual similarity clustering. 
Overall, moderate confusion neighborhoods provide the best balance between coverage and noise.

\section{Related Work}
\label{sec:related_work}
\subsection{Chinese Spelling and Query Correction}

\emph{Chinese spelling correction (CSC)} is typically studied on sentence-level text with richer contextual information, whereas \emph{Chinese query correction (CQC)} focuses on short search queries, where ambiguity is higher and vocabulary shift is faster. 
Because query correction directly affects retrieval and recommendation, CQC places stronger emphasis on intent preservation and control over editing behavior than general CSC.

Early correction methods often followed a noisy-channel formulation with candidate generation and ranking \citep{brill-moore-2000-improved}. 
More recent CSC systems increasingly adopt PLM-based ``detect-then-correct'' architectures, such as Soft-Masked BERT \citep{zhang-etal-2020-spelling}, to reduce unnecessary rewriting. 
For Chinese, phonetic and glyph similarity are especially important, and many models incorporate them through structured relations, multimodal features, including SpellGCN \citep{cheng-etal-2020-spellgcn} and ReaLiSe \citep{xu-etal-2021-read}. 
Other efficient correction systems, such as FASPell \citep{hong-etal-2019-faspell}, also show strong performance under realistic Chinese spelling noise, while newer datasets such as CSCD-NS broaden evaluation coverage \citep{hu-etal-2024-cscd}.

Within the query setting, recent work highlights the mismatch between generic CSC assumptions and real search queries. 
Benchmarks such as \textit{QSpell 250K} \citep{ye-etal-2025-qspell} better reflect naturally occurring query errors and rapidly changing entities. 
At the same time, recent studies show that strong language models can still over-correct short ambiguous inputs toward frequent or generic expressions \citep{li2023effectiveness,liu-etal-2024-relm}, motivating stronger mechanisms for edit control. 
To improve robustness on rare entities and shifting queries, some systems further incorporate retrieval or orchestration components, such as RACQC \citep{su-etal-2025-racqc} and Trigger$^{3}$ \citep{zhang-etal-2024-trigger3}.

Our work is related to these lines but differs in where the correction constraint is imposed. 
Prior CSC models typically inject phonetic or glyph knowledge into a supervised correction model, while training-free LLM approaches mainly constrain decoding at inference time. 
By contrast, \textsc{GUIDE} uses phonetic and visual similarity to define a shared-ID input space and learns correction through reconstruction from that lossy abstraction. 
This makes the confusion structure itself part of the training objective, rather than only an auxiliary feature or a decoding-time constraint.

\subsection{Unsupervised Query Correction}

To reduce reliance on annotated correction pairs, unsupervised and weakly supervised correction methods learn from raw text or logs via synthetic corruption, self-training, denoising, or masked reconstruction. 
A common strategy is to generate pseudo pairs by applying confusion-set substitutions or masking and then train a model to recover the original text \citep{liu-etal-2021-plome,li-2022-uchecker}. 
Several studies further argue that realistic Chinese spelling errors are strongly shaped by the input process, and therefore simulate pinyin IME decoding to create pseudo data that better matches real error patterns \citep{hu-etal-2024-cscd}. 
Other work reduces dependence on fixed confusion sets by using language-model scoring, self-supervised decoding, or inference-time denoising \citep{jiang-etal-2024-chinese,zhou-etal-2024-simple}. 
There are also flexible correction pipelines based on candidate evaluation with lightweight detectors and configurable candidate tables \citep{shao2023dual}.

Compared with these approaches, \textsc{GUIDE} takes a different view of unsupervised learning. 
Rather than constructing synthetic error--correction pairs, we learn directly from unlabeled query streams by mapping confusable characters into shared encoder IDs and training the model to reconstruct the original query. 
This confuse-then-clarify formulation avoids committing to a hand-designed corruption process while still imposing a strong inductive bias toward plausible IME-shaped edits. 
In this sense, \textsc{GUIDE} is closest in spirit to input-process-aware correction, but differs in using confusion neighborhoods as a lossy input abstraction and reconstruction signal for controlled generation.

\section{Limitations and Future Work}

\textsc{GUIDE} is designed for high-precision correction under weak query context, and this design choice also defines its current scope.

First, the framework focuses on length-preserving character substitution and does not directly address insertion, deletion, segmentation, or phrase-level rewriting errors.
This is a practical trade-off: in our production setting, substitution errors---especially homophonic and visually similar ones---are among the most practically important query mistakes, making controlled substitution correction highly valuable.
In practice, we view this as a deliberate decomposition choice: different query error types can be handled by the models or modules best suited to them, which is often a favorable design for industrial systems.

Second, although phonetic and visual shared-ID neighborhoods cover the dominant error sources in Chinese queries, they do not capture all realistic error types.
Errors falling outside the constructed neighborhoods, or cases where the neighborhood itself is noisy, may still lead to missed corrections or over-correction.
This is particularly relevant for visual clustering, whose neighborhoods can be substantially noisier than phonetic ones.
At the same time, our results validate the effectiveness of this neighborhood-based design. While the current construction is not perfect, it can be further improved through continued refinement of clustering methods, and its failure cases remain relatively interpretable.

Third, our current deployment combines phonetic and visual signals through a simple heuristic selection strategy rather than a unified fusion model.
We chose this decoupled design for stability and operational simplicity in production, but it leaves room for more principled integration, such as confidence-aware fusion, joint multimodal encoding, or learned routing between error types.

Finally, because our platform primarily serves Chinese queries, we focus this study on Chinese query correction.
That said, the overall confuse-then-clarify paradigm is not inherently Chinese-specific, and can be generalized to other languages with language-appropriate neighborhood construction, such as phonological neighbors, keyboard-proximity errors, orthographic variants, or morphology-aware confusion structures.
Exploring such extensions is a natural direction for future work.

\section{Conclusion}
\label{sec:conclusion}

We studied unsupervised Chinese query correction on content platforms, where weak context and IME-shaped noise make free-form rewriting especially prone to intent drift. 
\textsc{GUIDE} addresses this challenge through a confuse-then-clarify paradigm: by encoding queries in phonetic/visual shared-ID neighborhoods and reconstructing original character sequences, it enables controlled correction from unlabeled query streams. 
Across both offline benchmarks and online A/B testing, the results suggest that, for short-query correction, effective control over the edit space can be more important than unconstrained generation.

\section*{Acknowledgments}
This work was conducted at Kuaishou Technology. We thank the company for its support, and our colleagues on the search and
recommendation team for helpful discussions and their assistance with data processing, deployment, and online experiments. We are
especially grateful to Xuanping Li for valuable guidance and support throughout this work. We also thank the anonymous reviewers and area
chairs for their constructive comments.

\bibliography{custom}

 \clearpage 

\appendix
\section{Supplementary Experimental Settings}
\label{sec:appendix}

\subsection{Training Settings}
\label{app:training_settings}

This appendix provides the detailed training configurations for \textsc{GUIDE}. Unless otherwise specified, all reported results use the same training recipe across model sizes.

\paragraph{Hardware.}
We train \textsc{GUIDE} on a single machine equipped with $4\times$ NVIDIA L20 GPUs. Mixed-precision training is enabled.

\paragraph{Model Configuration.}
\textsc{GUIDE} uses a standard Transformer encoder--decoder architecture.
We evaluate three sizes with $\{3,6,12\}$ encoder layers and $\{3,6,12\}$ decoder layers, respectively.
For all variants, we use $\text{d\_model}=768$, $\text{d\_ff}=3072$, $\text{\#\ heads}=12$, $\text{dropout}=0.1$, and $\text{label smoothing}=0.1$.
The vocabulary consists of character IDs, and the encoder input is the shared-ID sequence produced by the clustering module (Section~\ref{sec:character_clustering}).

\paragraph{Optimization.}
We optimize the time-decayed, frequency-reweighted negative log-likelihood objective in Eq.~\ref{eq:time-decayed-nll}.
We use AdamW with $\beta_1=0.9$, $\beta_2=0.999$, $\epsilon=1\times10^{-5}$, $\text{weight decay}=0.01$, and gradient clipping norm=1.0.
The base learning rate is $5\times10^{-5}$, with a warmup of 5\% steps of the total steps followed by a cosine decay schedule (e.g., inverse-square-root / cosine decay).
Training runs for 2 epochs, with an effective batch size of 512 queries (micro-batch size 128 per GPU and gradient accumulation steps 4).
We set the maximum input length to 16 characters and truncate longer sequences.

\paragraph{Data and Reweighting.}
For each training example $(\vx_i,t_i)$, the shared-ID input $\tilde{\vx}_i$ is constructed by replacing each character with its phonetic/visual shared-ID (Section~\ref{sec:character_clustering}).
The weight is $w_i=\log(1+c_i)\cdot \exp(-\lambda\Delta t_i)$, where $\lambda=0.0098$ and $\Delta t_i$ is measured in $90$ days.
We sample training data proportionally to $w_i$, and update weights every week.

\paragraph{Decoding and Checkpoint Selection.}
At inference time, we use beam search with beam size 10, length penalty 1.0, and maximum decoding length $m_i$ (equal to the input length).
For the unified system, we run both phonetic-shared-ID and visual-shared-ID encodings and select the encoding whose best beam candidate yields the smaller score gap to the original query (Section~\ref{sec:chinese_query_correction}).
We select the final checkpoint using validation query-level F1 (or NLL) on a held-out set, evaluated every 2000 steps.


\subsection{Baseline Settings}
\label{app:baseline_settings}

For supervised and fine-tuning baselines, we follow the original implementations and report the results in Section~\ref{sec:experiments}. 
For \textsc{LLM-ICL}, we use the Qwen3 series (0.6B/1.7B/4B) under 0-shot and 10-shot settings with randomly sampled demonstrations, using the prompt templates shown below.

\paragraph{Prompt Templates (LLM-ICL).}\mbox{}\\
\begin{CJK*}{UTF8}{gkai}
\noindent\textbf{0-shot.}
\begin{quote}
你是一个query纠错专家。请针对输入query，判断是否包含错别字，并进行合理纠错。\\
\textnormal{Query: }\texttt{\{text\}}\\
\textnormal{Output: }
\end{quote}
\noindent\textit{English translation:} ``You are a query-correction expert. For the input query, decide whether it contains misspelled
characters and correct them appropriately.''

\noindent\textbf{10-shot.}
\begin{quote}
你是一个query纠错专家。请针对输入query，进行合理的文本纠错，输出纠错后文本，禁止输出任何思考过程。\\
示例：\\
$\left.
\begin{array}{@{}l@{}}
\textnormal{Query: }\texttt{...}\\
\textnormal{Output: }\texttt{...}
\end{array}
\right\}\times 10$\\
\textnormal{Query: }\texttt{\{text\}}\\
\textnormal{Output: }
\end{quote}
\noindent\textit{English translation:} ``You are a query-correction expert. For the input query, perform reasonable text correction and
  output only the corrected text; do not output any reasoning process.'' \textnormal{(``示例'' = ``Examples''.)}
\end{CJK*}

 \section{Additional Experimental Results}
  \label{app:additional_results}

  \subsection{Ablation on the Training Objective and Sensitivity to $\lambda$}
  \label{app:reweighting_ablation}

  We ablate the two components of our training objective in Eq.~(\ref{eq:time-decayed-nll})  on \textit{KwaiSearch} to isolate their contributions. Starting from a uniform
  objective, adding query-frequency reweighting alone improves F1 by about 11 points, and further adding the time-decay term yields an
  additional gain of about 4.6 points (Table~\ref{tab:objective_ablation}). This confirms that frequency reweighting is the dominant
  factor, while time decay provides a consistent additional benefit by shifting learning focus toward recent queries.

  Importantly, $\lambda$ is not tuned on the test set. We instead set it from a simple operational assumption: with an approximately 60\%
  one-year query overlap and weekly model updates, requiring $e^{-\lambda\cdot 52}=0.6$ over the 52 weekly steps of a year gives $
  \lambda=0.0098$, which is the value used throughout our experiments. To show that this choice is robust, we also report more aggressive
  and more conservative alternatives obtained by targeting a 50\% overlap ($\lambda=0.0133$) or an 80\% overlap ($\lambda=0.0043$). As
  shown in Table~\ref{tab:lambda_sensitivity}, both variants keep F1 within about 1.6 points of our default, indicating that \textsc{GUIDE}
  is not sensitive to the exact value of $\lambda$.

  \begin{table}[t]
  \centering
  \small
  \setlength{\tabcolsep}{5pt}
  \begin{tabular}{lccc}
  \toprule
  \textbf{Objective} & \textbf{Prec.} & \textbf{Rec.} & \textbf{F1} \\
  \midrule
  Uniform & 0.5503 & 0.6470 & 0.5947 \\
  Frequency-only & 0.6314 & 0.7974 & 0.7048 \\
  Full (time-decay + freq.) & \textbf{0.6738} & \textbf{0.8476} & \textbf{0.7508} \\
  \bottomrule
  \end{tabular}
  \caption{Ablation of the training objective on \textit{Kwaisearch} (3-layer model). Frequency reweighting alone adds $\sim$11 F1 points
  over the uniform objective, and time decay adds a further $\sim$4.6 points.}
  \label{tab:objective_ablation}
  \end{table}

  \begin{table}[t]
  \centering
  \small
  \setlength{\tabcolsep}{5pt}
  \begin{tabular}{lccc}
  \toprule
  \textbf{$\lambda$} & \textbf{Prec.} & \textbf{Rec.} & \textbf{F1} \\
  \midrule
  0.0098 (ours) & 0.6738 & \textbf{0.8476} & \textbf{0.7508} \\
  0.0133 (aggressive) & 0.6680 & 0.8422 & 0.7451 \\
  0.0043 (conservative) & \textbf{0.6817} & 0.7974 & 0.7350 \\
  \bottomrule
  \end{tabular}
  \caption{Sensitivity to the time-decay coefficient $\lambda$ on \textit{Kwaisearch} (3-layer model). Targeting 50\% overlap ($0.0133$) or
  80\% overlap ($0.0043$) keeps F1 within $\sim$1.6 points of our default $\lambda=0.0098$.}
  \label{tab:lambda_sensitivity}
  \end{table}

  \subsection{Selection of the Visual Clustering Threshold $\tau$}
  \label{app:tau_selection}

  The visual clustering threshold $\tau$ is selected on a separate held-out development set consisting purely of visual errors, rather than
  on the \textit{Kwaisearch} test set. We sweep $\tau$ for the visual-similarity model on this development set and report the results in
  Table~\ref{tab:tau_selection}. The threshold $\tau=0.5$ is optimal on the development set, and we keep this value throughout all
  experiments. This makes the selection protocol explicit and ensures that $\tau$ is not tuned on any test data.

  \begin{table}[t]
  \centering
  \small
  \setlength{\tabcolsep}{5pt}
  \begin{tabular}{lccc}
  \toprule
  \textbf{Threshold $\tau$} & \textbf{Prec.} & \textbf{Rec.} & \textbf{F1} \\
  \midrule
  0.3 & 0.5974 & 0.7613 & 0.6695 \\
  0.4 & 0.5969 & 0.7774 & 0.6753 \\
  0.5 (selected) & \textbf{0.5999} & \textbf{0.7890} & \textbf{0.6816} \\
  0.6 & 0.5974 & 0.7347 & 0.6590 \\
  \bottomrule
  \end{tabular}
  \caption{Selection of the visual clustering threshold $\tau$ on a held-out development set of visual errors only. $\tau=0.5$ is optimal
  and is used throughout.}
  \label{tab:tau_selection}
  \end{table}

\subsection{Case Studies}
\label{app:case_study}

We provide qualitative examples to complement the quantitative results in Section~\ref{sec:experiments}. 
Table~\ref{tab:case_study} includes both representative \emph{good cases} and \emph{bad cases} for \textsc{GUIDE}. 
In good cases, \textsc{GUIDE} fixes typical IME-shaped confusions (sound-alike or look-alike) while preserving the remaining characters, producing outputs that better match the intended query. 
In bad cases, two common failure modes are observed: (i) \emph{over-correction}, where a rare but valid query is rewritten toward a more frequent alternative, and (ii) \emph{missed correction}, where the error lies outside the clustering neighborhood or requires extra context beyond the query itself. 
\begin{CJK*}{UTF8}{gkai}
For instance, \textsc{GUIDE} corrects ``时间同步板远离'' $\rightarrow$ ``时间同步板原理'' and ``臧易通怎么查核酸报告'' $\rightarrow$ ``藏易通怎么查核酸报告'', but may over-correct ``播音苏杨'' to the frequent phrase ``播音素养'', or keep ``name英标怎么写'' unchanged when the cue is weak; this mainly happens because short queries provide little context, so the model tends to favor high-probability frequent phrases unless the phonetic/visual neighborhood offers a strong and unambiguous correction signal.
\end{CJK*}

\begin{CJK*}{UTF8}{gkai}
\begin{table}[t]
\centering
\small
\setlength{\tabcolsep}{3pt}
\begin{tabular}{l|p{2.4cm}|p{3.3cm}}
\toprule
\textbf{Case} & \textbf{Good (Pho)} & \textbf{Good (Vis)} \\
\midrule
\textbf{Original} & 时间同步板远离 & 臧易通怎么查核酸报告 \\
\textbf{Target}   & 时间同步板原理 & 藏易通怎么查核酸报告 \\
\midrule
\textsc{GUIDE}      & 时间同步板原理 & 藏易通怎么查核酸报告 \\
\textsc{BERT}       & 时间同步板远离 & 交易通怎么查核酸报 \\
\textsc{Masked-FT}  & 时间同步被远离 & 易易通怎么查核酸报告 \\
\textsc{Simple-CSC} & 时间同步板远离 & 臧易通怎么查核酸报告 \\
\textsc{LLM-ICL}    & 时间同步板远离 & 臧易通如何查询核酸报告 \\
\midrule
\textbf{Case} & \textbf{Bad (Over)} & \textbf{Bad (Miss)} \\
\midrule
\textbf{Original} & 播音苏杨 & name英标怎么写 \\
\textbf{Target}   & 播音苏杨 & name音标怎么写 \\
\midrule
\textsc{GUIDE}      & 播音素养 & name英标怎么写 \\
\textsc{BERT}       & 播音苏杨 & name英标怎么写 \\
\textsc{Masked-FT}  & 播音苏扬 & name英标怎么写 \\
\textsc{Simple-CSC} & 播音苏扬 & name英标怎么写 \\
\textsc{LLM-ICL}    & 播音苏扬 & name的英标是/nɛm/ \\
\bottomrule
\end{tabular}
\caption{Qualitative case studies on representative query correction examples. We show the original query, the target correction, and the outputs of \textsc{GUIDE} and baseline methods for two good cases (Pho/Vis) and two failure cases (Over/Miss).}
\label{tab:case_study}
\vspace{-1em}
\end{table}
\end{CJK*}

\begin{table}[t]
\centering
\small
\setlength{\tabcolsep}{5pt}
\begin{tabular}{lccc}
\toprule
\textbf{Setting} & \textbf{Prec.} & \textbf{Rec.} & \textbf{F1} \\
\midrule
\multicolumn{4}{l}{\textit{Phonetic clustering}} \\
Coarse pinyin clusters & 0.6020 & 0.7650 & 0.6738 \\
Default homophonic clustering & \textbf{0.6662} & \textbf{0.8433} & \textbf{0.7443} \\
Tone-aware clustering  & 0.4585 & 0.8266 & 0.5898 \\
\midrule
\multicolumn{4}{l}{\textit{Visual clustering threshold}} \\
$\tau=0.3$ & 0.1245& 0.7859& 0.2149
\\
$\tau=0.4$ & 0.1267& 0.7909& 0.2184
\\
$\tau=0.5$ & \textbf{0.1326}& \textbf{0.7971}& \textbf{0.2274}
\\
$\tau=0.6$ & 0.1311& 0.7938& 0.2250\\
\bottomrule
\end{tabular}
\caption{Full numerical results for the clustering-granularity analysis on \textit{KwaiSearch}.}
\label{tab:granularity_sensitivity}
\vspace{-1em}
\end{table}

\section{Online A/B Testing Details}
  \label{app:ab_details}

  The production baseline in our online A/B test (Section~\ref{subsec:AB_testing}) is an LLM-based corrector: a Qwen model fine-tuned on tens
  of thousands of manually annotated correction pairs, using the same prompt as provided in this paper
  (Appendix~\ref{app:baseline_settings}). We compare \textsc{GUIDE} against this deployed LLM-based correction model.

  We run a bucketed A/B test from 2025-09-13 to 2025-09-22 (10 days) on 4.2\% of production main-search traffic. Our single-day search page
  views (PV) are on the order of tens of millions, and the downstream query-recommendation scenarios fed by the corrector carry far larger
  PV than search itself; at this scale, the A/B conclusions are highly confident. The misspelling rate is measured by random sampling of
  online queries followed by manual annotation. During the bucketed test, \textsc{GUIDE} reduces the misspelling rate from 2.58\% (base) to
  0.86\% (exp), and yields a $+0.122\%$ overall search-volume lift.

  \textsc{GUIDE} was fully rolled out at the end of September 2025. We continue to monitor the online misspelling rate by month, as
  reported in Table~\ref{tab:ab_tracking}. The reduction remains stable across subsequent months rather than being a transient effect,
  confirming a sustained improvement in correction quality.

  \begin{table}[t]
  \centering
  \small
  \setlength{\tabcolsep}{6pt}
  \begin{tabular}{lcccc}
  \toprule
  \textbf{Month} & 09-01 & 10-01 & 11-01 & 12-01 \\
  \midrule
  \textbf{Misspelling rate} & 2.01\% & 0.40\% & 0.46\% & 0.33\% \\
  \bottomrule
  \end{tabular}
  \caption{Post-rollout monthly tracking of the online misspelling rate after the full rollout of \textsc{GUIDE} (end of September 2025).
  The reduction is stable and sustained.}
  \label{tab:ab_tracking}
  \end{table}

\end{document}